\pdfoutput=1

\documentclass[11pt]{article}

\usepackage{EACL2023}

\usepackage{times}
\usepackage{latexsym}

\usepackage[T1]{fontenc}

\usepackage[utf8]{inputenc}

\usepackage{microtype}

\usepackage{inconsolata}

\usepackage{color}
\usepackage{soul}
\usepackage{amsmath}
\usepackage{amsfonts}
\usepackage{amssymb}
\usepackage{pifont}
\usepackage{xcolor}
\usepackage{graphicx}
\usepackage{enumitem}
\usepackage{booktabs}
\usepackage{multirow}
\usepackage{array}
\usepackage{float}
\usepackage{arydshln}
\usepackage{adjustbox}
\usepackage{multicol}
\usepackage{mathabx}
\usepackage{soul}
\usepackage[section]{placeins}
\usepackage[normalem]{ulem}

\usepackage[multiple]{footmisc}

\setlist[itemize]{itemsep=0.02cm,topsep=0.2cm}

\newcommand{\cmark}{\ding{51}}%
\newcommand{\xmark}{\ding{55}}%

\title{Mind the Labels: Describing Relations in Knowledge Graphs With Pretrained Models}

\def\cuni{\textsuperscript{1}}
\def\hwu{\textsuperscript{2}}

\author{Zdeněk Kasner,\cuni\ \ Ioannis Konstas\hwu \and Ondřej Dušek\cuni \\
  \cuni{}Charles University, Faculty of Mathematics and Physics, Prague, Czechia \\
  \hwu{}The Interaction Lab, MACS, Heriot-Watt University, Edinburgh, UK  \\
  \texttt{\{kasner,odusek\}@ufal.mff.cuni.cz, i.konstas@hw.ac.uk}
}

\usepackage{xcolor}
\usepackage[normalem]{ulem}

\definecolor{lightblue}{RGB}{115, 193, 214}

\definecolor{lightyellow}{RGB}{255, 230, 128}

\definecolor{grey}{RGB}{66, 66, 66}
\newcommand{\grey}[1]{{\leavevmode\color{grey}{#1}}}
\newcommand{\green}[1]{{\leavevmode\color{green!60!black!100}{#1}}}
\newcommand{\red}[1]{{\leavevmode\color{red!60!black!100}{#1}}}

\newcommand{\eh}{\sethlcolor{lightblue}\grey{\hl{X}}}
\newcommand{\et}{\sethlcolor{lightyellow}\grey{\hl{Y}}}

\newcommand{\BARTr}{\textit{full-rel2text}}
\newcommand{\BARTk}{\textit{full-kelm}}
\newcommand{\BARTw}{\textit{full-webnlg}}

\newcommand\Tstrut{\rule{0pt}{2.2ex}}
\newcommand\Bstrut{\rule[-0.6ex]{0pt}{0pt}}

\begin{document}
\maketitle
\begin{abstract}
  Pretrained language models (PLMs) for data-to-text (D2T) generation can use \textit{human-readable data labels} such as column headings, keys, or relation names to generalize to out-of-domain examples.
  However, the models are well-known in producing semantically inaccurate outputs if these labels are ambiguous or incomplete, which is often the case in D2T datasets. In this paper, we expose this issue on the task of descibing a relation between two entities. For our experiments, we collect a novel dataset for verbalizing a diverse set of 1,522 unique relations from three large-scale knowledge graphs (Wikidata, DBPedia, YAGO). We find that although PLMs for D2T generation expectedly fail on unclear cases, models trained with a large variety of relation labels are surprisingly robust in verbalizing novel, unseen relations. We argue that using data with a diverse set of clear and meaningful labels is key to training D2T generation systems capable of generalizing to novel domains.\footnote{We release the code and data for our experiments: \url{https://github.com/kasnerz/rel2text}.}
\end{abstract}

\section{Introduction}
D2T generation systems need to accurately capture the semantics of relations between values in the data. However, the data labels such as relation names \cite{farber2018linked,haller2022analysis}, table headings \cite{parikh2020totto}, or meaning representation keys \cite{duvsek2020evaluating_challenge} may provide only superficial or---if the labels are abbreviations, such as in the Rotowire dataset \cite{wiseman2017challenges}---no usable hints about the data semantics. Learning how to properly describe the data is thus a challenge for D2T systems, typically requiring in-domain training data of sufficient quality and quantity \cite{duvsek2019semantic}.

\begin{figure}[t]
  \centering
  \includegraphics[width=\columnwidth]{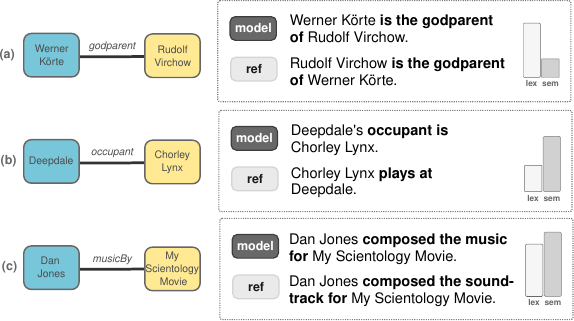}
  \caption{Data-to-text generation models use relation labels (such as  \mbox{\textit{godparent}}, \mbox{\textit{occupant}}, and \mbox{\textit{musicBy}}) to describe relations between entities. However, unclear labels can lead to various lexical or semantic incoherencies in the output descriptions, such as swapping the relation direction (a) or using too literal expressions (b).}\label{fig:teaser}
\end{figure}

PLMs such as BART \cite{lewis2020bart} or T5 \cite{raffel2020exploring} can quickly adapt to new domains and exhibit robustness to out-of-domain inputs. However, the PLMs for D2T generation are still limited by the expressivity of the data labels. Consider Figure~\ref{fig:teaser} (a): the model can use its representation of \textit{``godparent''} to understand there is a \textit{``is-a-godparent-of''} relation between the entities, but it has to infer (or guess) who is the godparent of whom. Even in the less ambiguous cases (b) and (c), the model still has to correctly capture the intended semantics of the relation (e.g. \textit{``occupant''} meaning \textit{``home team''}).

In this paper, we investigate to what extent PLMs are able to use arbitrary labels describing relations between entities. A suitable testing ground is the task of describing (i.e., \textit{verbalizing}) individual triples in a knowledge graph (KG), which can be considered a trivial case of graph-to-text (G2T) generation \cite{ribeiro2020investigating,koncel2019text}. In this task, there is a wide range of lexical choices for the \textit{relation label} (see Table \ref{tab:example}), while the \textit{entities} can be copied verbatim or with only minor morphological changes.

Current human-annotated datasets for D2T generation contain only a small number of relations
and rarely contain any unseen relations in the test set \cite{mille2021automatic}.
We collect a novel dataset \textsc{Rel2Text} (\underline{R}e-writing \underline{e}dge \underline{l}abels to Text),\footnote{Or simply ``Relations-to-Text''.} acting as a test bench for our experiments. It contains 4,097 single triples from three large-scale KGs (Wikidata, \mbox{DBPedia}, and YAGO) and their crowdsourced verbalizations, covering 1,522 unique relations (§\ref{sec:data}). Each relation is equipped with a label, a textual description, and up to five triples in which the relation occurs in the KG.

Using the \textsc{Rel2Text} dataset, we evalute the ability of PLMs to verbalize relations which were not present in the training set. We consider both models finetuned on other relations in our dataset and models finetuned on datasets from a related domain. We also experiment with scenarios involving few-shot finetuning, training on masked labels, and extending the labels with descriptions (§\ref{sec:analysis},~\ref{sec:results}).

We find that the PLMs are quite robust in verbalizing a diverse set of relations based on their label (achieving \textasciitilde 90\% of overall entailment probability). We show that semantically unfaithful model outputs are often caused by incomplete, ambiguous, or noisy input data. Somewhat suprisingly, we also show that longer relation descriptions do not provide substantial improvements over using short labels. However, even for data using short relation labels, the model trained on verbalizing relations can achieve results comparable to verbalizing relations using manual templates in two downstream tasks (§\ref{sec:downstream}).

\begin{table}[t!] \footnotesize
  \def\arraystretch{1.2}
  \centering\begin{tabular}{lp{5.3cm}} \toprule
    \textbf{relation}   & \textbf{possible verbalization}                                      \\ \midrule
    \textit{is part of} & \eh{} is part of \et{}.                                              \\\hdashline[0.5pt/2pt]
    \textit{duration}   & \eh{} lasted for \et{}.                                              \\\hdashline[0.5pt/2pt]
    \textit{platform}   & \eh{} is available on \et{}.\newline\eh{} runs on \et{}.             \\\hdashline[0.5pt/2pt]
    \textit{country}    & \eh{} was born in \et{}. \newline \eh{} is located in \et{}.         \\\hdashline[0.5pt/2pt]
    \textit{parent}     & \eh{} is the parent of \et{}. \newline \et{} is the parent of \eh{}. \\\hdashline[0.5pt/2pt]
    \textit{ChEMBL}     & \eh{} has an id \et{} in the ChEMBL database.                        \\ \bottomrule
  \end{tabular}
  \caption{Examples of relation labels and their possible verbalizations, with placeholders for head (\eh) and tail (\et) entities. 
    Relations can be copied verbatim (\textit{is part of}), have a unique verbalization (\textit{duration}), or multiple equivalent lexical choices (\textit{platform}). There 
    is also ambiguity stemming from the semantics of the entities (\textit{country}) or the relation itself (\textit{parent}, \textit{ChEMBL}).}
  \label{tab:example}
\end{table}

The contributions of our work are as follows:
\begin{itemize}
  \item We examine the ability of PLMs to describe graph relations, showing that \textit{clear and meaningful labels} are the basis for successful generalization to unseen relations.
  \item We present \textsc{Rel2Text}---a human-annotated dataset with 4,097 examples verbalizing 1,522 relations from three large-scale open KGs.
  \item We show that a model trained on \textsc{Rel2Text} can serve as a drop-in replacement for manual templates, preserving or improving performance on downstream tasks.
\end{itemize}

\section{Related Work}
\label{sec:rel_work}

Earlier works in natural language generation from KGs exploited domain-specific ontologies for rule-based systems \cite{cimiano2013exploiting,bouayad2012ontology,sun2007experiment,sun2006domain}. With the advance of PLMs, structure-aware modeling and task-specific pretraining has lead to remarkable progress on \textbf{D2T benchmarks} such as WebNLG \cite{gardent2017webnlg,ferreira20202020}, AGENDA \cite{koncel2019text}, or E2E \cite{duvsek2020evaluating_challenge}, indicated via both automatic and human evaluation metrics \cite{ke2021jointgt,guo2020cyclegt,ribeiro2020modeling,harkous2020have}.

The existing datasets covering verbalizations of a wider range of KG relations are based on \textbf{model-generated outputs}. \citet{agarwal2021knowledge} used semantic filtering and distant supervision for generating a large-scale corpus of synthetic verbalizations of the English Wikidata. We use their KeLM corpus to investigate how training on large-scale synthetic data differs from training on a small-scale human-annotated dataset (see\ §\ref{sec:analysis}). Parallel to us, \citet{amaral2022wdv} introduced the WDV dataset with verbalizations of individual Wikidata triples. The scope and size of their dataset is similar to ours, but their verbalizations are human-validated outputs of the T5 model \cite{raffel2020exploring} finetuned on the WebNLG dataset.

Other works have tried \textbf{incorporating descriptions of data labels} in the model inputs. In one of the experiments, \citet{wang2021kepler} use descriptions of relations from Wikidata instead of their labels for relation embeddings, concluding that it results in worse performance on downstream tasks. Conversely, \citet{kale-rastogi-2020-template} and \citet{lee2021dialogue} improve the performance of their systems by including schema descriptions on the input for the dialogue state tracking and dialogue response generation systems.

There has also been a research interest in \textbf{verbalizing single triples} as a stand-alone preprocessing step for NLP tasks. The step has been shown to improve the generalization ability of downstream models for data-to-text generation \cite{laha2020scalable,kasner2020data,kasner2022neural,xiang2022asdot} and response generation in dialogue systems \cite{kale-rastogi-2020-template}. This step can also serve for making the input similar to the format used during pretraining, e.g.\ for  natural language inference (NLI) models (\citealp{gupta2020infotabs,neeraja2021incorporating,duvsek2020evaluating}). The above works employ a variety of methods to convert triples to text, ranging from simple templates and rule-based systems to prompting large PLMs. However, none of these works investigate how PLMs behave when presented with novel relations.

In a work concurrent to ours, \citet{keymanesh2022makes} investigate the aspects of \textbf{generalization performance of PLMs} on the DART dataset\footnote{We did not use DART (which is a compilation of several datasets including WebNLG) for our experiments since it contains many noisy relations.} \cite{nan2021dart}. They compare prompt-based and finetuning-based approaches to D2T generation, focusing on the ability of models to perform on difficult examples. In contrast, we focus on finetuned encoder-decoder models, which were shown in \citet{keymanesh2022makes} to be more efficient for D2T generation, and we evaluate the models on clean and manually curated data.

\section{Data}
\label{sec:data}
For our experiments, we need data with diverse labels and their human verbalizations. In this section, we describe how we gather RDF\footnote{\url{https://www.w3.org/TR/PR-rdf-syntax/}} triples from large-scale KGs (§\ref{sec:input_data}) and collect their verbalization through crowdsourcing (§\ref{sec:rel2text},\ \ref{sec:postprocessing}).

\subsection{Input Data}
\label{sec:input_data}
An RDF triple is a tuple $t = (e_h, r, e_t)$, where $r$ denotes the relation between the head entity $e_h$ and the tail entity $e_t$.
We retrieve triples from three open large-scale KGs encoding factual knowledge:

\begin{itemize}
  \item \textbf{Wikidata} \cite{vrandevcic2014wikidata} is a large-scale Wikipedia-based KG created using collaborative editing. With approximately 10,000 human-created relations equipped with descriptions,\footnote{\url{https://www.wikidata.org/wiki/Wikidata:Database_reports/List_of_properties/all}} it is by far the largest source of variety in relation labels.
  \item \textbf{YAGO} \cite{pellissier2020yago} is a KG which builds upon factual knowledge from Wikidata, but uses a limited set of 116 pre-defined relations from \texttt{schema.org} \cite{guha2016schema} mapped to a subset of Wikidata relations.
  \item \textbf{DBPedia} \cite{lehmann2015dbpedia} is a KG that maps Wikipedia infotables to a predefined ontology containing 1,355 relations, about 350 of which are accompanied by a description.
\end{itemize}

We query all KGs using their openly available endpoints to retrieve a list of relations in each KG. For each relation, we retrieve up to five \textit{triples} that use this relation, and the relation \textit{description}, i.e.\ a short explanatory text.
If present, we also retrieve descriptions for the head and tail entities.

We apply a set of filtering heuristics, leaving out e.g.\ relations describing KG metadata or identification numbers.\footnote{Relations describing various IDs make up a suprisingly large portion of relations in Wikidata. Since we focus on diversity instead of coverage, we decided not to include these relations in our dataset.} In this way, we collect 7,334 triples with 1,716 relations in total. For the full description regarding the data retrieval, please refer to Appendix~\ref{app:scraping}.

\subsection{Annotation Process}
\label{sec:rel2text}
We collect human-written verbalizations for all input triples using Prolific.\footnote{\url{https://www.prolific.co/}} We built a web interface in which the human annotators are shown a single triple $t$ and asked to describe it in a single sentence. The annotators are encouraged to re-use the entities in their original form, but they are able to change the form if necessary. The annotators can also report noisy inputs. We employed 420 annotators in total, each of which annotated 20 examples. We set the average reward per hour according to the platform recommendations to  \pounds 7.29 per hour and we accepted all the inputs which passed our built-in checks. See Appendix \ref{app:crowdsourcing} for more details on the annotation process.

\subsection{Postprocessing the Data}
\label{sec:postprocessing}
A considerable portion of the collected verbalizations contain typos and grammatical errors, misunderstood meaning of the relation, or extra information in the input. To ensure high quality of our data, we manually examined all crowdsourced examples and annotated them as \textit{OK}, \textit{noisy}, \textit{corrupted} or \textit{containing extra information}.
Appendix \ref{app:postprocessing} includes postprocessing details.
In the rest of the paper, we only use the subset of our dataset with \textit{OK} annotations, one per input triple (4,097 examples, 1,522 distinct relations), although we also make the remaining noisy instances available for future research.

\section{Analysis and Evaluation}
\label{sec:analysis}
In our analysis, we are interested in the following research questions:
\begin{itemize}
  \item \textbf{RQ1:} Are the PLMs finetuned for D2T generation able to describe relations \textit{not present in the finetuning corpus}?
  \item \textbf{RQ2:} How many \textit{training examples} do the PLMs need to generate satisfactory outputs?
  \item \textbf{RQ3:} How do the PLMs behave when provided \textit{limited lexical cues} about the relation?
  \item \textbf{RQ4:} Can relation \textit{descriptions} help to clarify ambiguous cases and improve semantic accuracy of the outputs?
\end{itemize}

To answer these questions, we divide our \textsc{Rel2text} dataset into a training and test splits (see §\ref{sec:setup} for details). We then use the \textbf{\textsc{Rel2Text} test set} to evaluate a finetuned BART model \cite{lewis2020bart}, a pretrained encoder-decoder transformer, which is used as a backbone of many recent data-to-text models (\citealp{ke2021jointgt,xing2021structure,ribeiro2020investigating,liu2021kg}).\footnote{We believe that our findings also apply to similar models such as T5 \cite{raffel2020exploring}, which have shown comparable performance on related tasks.}

To answer \textit{RQ1}, we compare the performance of BART finetuned on the \textsc{Rel2Text} training set with BART finetuned on two qualitatively different D2T datasets -- \textsc{WebNLG} and \textsc{KeLM}. Using \textsc{Rel2text} only, we then prepare various setups for answering \textit{RQ2}, \textit{RQ3}, and \textit{RQ4} (details in §\ref{sec:experiments}). We analyze the outputs of the models  both automatically (§\ref{sec:auto}) and manually (§\ref{sec:manual}).

\subsection{Experimental Setup}
\label{sec:setup}

\paragraph*{Datasets}  We experiment with the following datasets, all of which focus on verbalizing factual information from KGs and use the same triple-based input data format:
\begin{itemize}
  \item \textsc{Rel2Text}. Our dataset (cf.\ §\ref{sec:rel2text}) with single triples from three KGs with 4,097 examples, 1,522 relations and \textit{human-annotated} outputs.
  \item \textsc{WebNLG} \cite{ferreira20202020,gardent2017webnlg}. A DBPedia-based triple-to-text dataset with 38k examples, 411 relations, up to 7 triples per example, and \textit{human-annotated} outputs. We use the English part of version 3.0 from HuggingFace.\footnote{\url{https://huggingface.co/datasets/web_nlg}}
  \item \textsc{KeLM} \cite{agarwal2021knowledge}. A Wikidata-based dataset with 11M examples, 1,519 relations, up to 13 triples per example, and \textit{model-generated} outputs. We use the dataset released by the authors, splitting it in a 1:100 ratio into validation and training data.
\end{itemize}

\paragraph{Rel2Text Data Split} We use approximately 15\% of the \textsc{Rel2Text} examples for the \textbf{test set}. To ensure maximum fairness and focus on model generalization to unseen relations, we do not include in the \textsc{Rel2Text} test set any relations which have an exact string match with a relation in \textsc{KeLM}, \textsc{WebNLG}, or the \textsc{Rel2Text} training set. We also exclude any relations for which the maximum semantic similarity\footnote{Computed as cosine similarity between embeddings of the labels, which are encoded using \texttt{all-distilroberta-v1} from SBERT \cite{reimers-2019-sentence-bert}.} to any \textsc{KeLM}/\textsc{WebNLG}/\textsc{Rel2Text} training relation exceeds a threshold of $0.9$. We set this threshold empirically in order to exclude relations which are almost synonymous, but slightly lexically different.
We use 90\% of the remaining examples for the training set and 10\% for the validation set.

\paragraph{Data Preprocessing} We split the camel case in the relation labels. For finetuning the models, we linearize the input triples by marking the triple constituents with special tokens \textit{<head>}, \textit{<rel>} and \textit{<tail>}, which we add to the model vocabulary.

\paragraph{Training and Decoding Setup} In a default scenario, we finetune BART\textsc{-base} for 10 epochs and select the best checkpoint using validation BLEU score, then use greedy decoding to produce outputs. We repeat each experiment with five random seeds, averaging the results. See Appendix~\ref{app:hyperparams} for details.

\subsection{Compared Systems}
\label{sec:experiments}
\paragraph{Copy Baseline} We introduce a simple baseline by outputting the triple constituents separated by space: ``$e_h\text{ }r\text{ }e_t$''.

\paragraph{Full Training Data} We use the default setup (§\ref{sec:setup}) on full \textsc{Rel2Text} and \textsc{WebNLG} training sets. For \textsc{KeLM} (which is about 300$\times$ larger than WebNLG), we finetune the model for 1 epoch only. We denote the trained models \BARTr{}, \BARTw{}, and \BARTk{}, respectively.

\paragraph{Limited Training Data} For the limited training data setup, we prepare few-shot splits from \textsc{Rel2Text} as subsets containing $N=$ \{25, 50, 100, 200\} relations with a single example per relation. We select examples at random, ensuring that each few-shot split is a subset of the larger splits. We finetune the \textit{fewshot-N} models for 10 epochs without validation, using the last checkpoint.

\paragraph{Limited Lexical Cues} In D2T datasets (with certain exceptions, cf.\ \citet{gardent2017creating}), unclear labels are kept in original form, implicitly assuming that the models will learn the verbalizations from the training data. We investigate how the models behave if we take this issue to the extreme, i.e. if the relation labels are not available at all. We consider three scenarios:
\begin{itemize}
  \item \textit{mask-test} -- We train the model on \textsc{Rel2Text} in the standard training setup. For testing, we replace the relation labels in  \textsc{Rel2Text} with the \textit{<mask>} token.
  \item \textit{mask-train} -- For training, we replace the relation labels in  \textsc{Rel2Text} with the \textit{<mask>} token. We test the model on \textsc{Rel2Text} in the standard evaluation setup.
  \item \textit{mask-all} -- We replace the relation labels in  \textsc{Rel2Text} with the \textit{<mask>} token for both training and testing.
\end{itemize}

\paragraph{Incorporating Descriptions} Our dataset contains short textual descriptions of the relations, which may be useful to disambiguate its meaning and provide additional cues to the model. We consider two scenarios:
\begin{itemize}
  \item \textit{desc-repl} -- We replace the relation label with its description.
  \item \textit{desc-cat} -- We concatenate the relation description with the input, separated using the special token \textit{<rel\_desc>}.
\end{itemize}

\begin{table*}[!htp]\centering
  \footnotesize
  \setlength{\tabcolsep}{5pt}
  \begin{tabular}{lc>{\hspace{-2mm}}c>{\hspace{-2mm}}c>{\hspace{-2mm}}ccccccc>{\hspace{-2mm}}c>{\hspace{-2mm}}c>{\hspace{-1mm}}cc}\toprule
    \multirow{2}{*}{} & \multicolumn{3}{c}{\textbf{Lexical}} & \multicolumn{5}{c}{\textbf{Semantics}} & \multicolumn{5}{c}{\textbf{Referenceless}}                                                                                                     \\\cmidrule(r){2-4}\cmidrule(r){5-9}\cmidrule{10-14}
                      & \bf BLEU                             & \bf METEOR                             & \bf BLEURT                                 & \bf SS & \bf C & \bf N & \bf E & \bf NB & \bf U-1 & \bf CE-2 & \bf MSTTR & \bf PPL & \bf len      \\\midrule
    \it human         & -                                    & -                                      & -                                          & -      & -     & -     & -     & -      & 1785    & 2.13     & 0.62      & 5.88    & 9.55         \\
    \it copy          & 29.04                                & 37.52                                  & 0.09                                       & 4.79   & 1.22  & 7.57  & 91.21 & 0.74   & 1606    & 1.17     & 0.7       & 7.55    & 6.72\Bstrut  \\\hdashline[0.5pt/2pt]
    \it \BARTr{}      & 52.54                                & 44.86                                  & 0.54                                       & 4.72   & 3.50  & 4.65  & 91.85 & 0.88   & 1661    & 1.96     & 0.58      & 5.89    & 9.16\Tstrut  \\
    \it \BARTw{}      & 41.99                                & 41.59                                  & 0.41                                       & 4.65   & 3.68  & 6.93  & 89.39 & 0.86   & 1651    & 2.54     & 0.56      & 5.65    & 10.29        \\
    \it \BARTk{}      & 46.74                                & 42.94                                  & 0.46                                       & 4.70   & 3.95  & 5.29  & 90.77 & 0.86   & 1652    & 2.32     & 0.56      & 5.83    & 9.71\Bstrut  \\\hdashline[0.5pt/2pt]
    \it fewshot-25    & 31.13                                & 35.52                                  & -0.02                                      & 3.94   & 8.35  & 27.26 & 64.39 & 0.65   & 1445    & 2.93     & 0.52      & 5.34    & 10.67\Tstrut \\
    \it fewshot-50    & 40.60                                & 40.05                                  & 0.25                                       & 4.44   & 8.04  & 13.12 & 78.84 & 0.76   & 1536    & 2.31     & 0.55      & 5.79    & 9.90         \\
    \it fewshot-100   & 45.88                                & 42.38                                  & 0.38                                       & 4.53   & 6.34  & 10.60 & 83.06 & 0.81   & 1600    & 2.13     & 0.57      & 5.85    & 9.57         \\
    \it fewshot-200   & 48.67                                & 43.34                                  & 0.44                                       & 4.58   & 5.40  & 9.03  & 85.57 & 0.83   & 1626    & 2.04     & 0.58      & 5.89    & 9.36\Bstrut  \\\hdashline[0.5pt/2pt]
    \it mask-test     & 42.45                                & 38.52                                  & 0.25                                       & 3.99   & 14.91 & 18.47 & 66.62 & 0.65   & 1669    & 1.96     & 0.61      & 5.69    & 8.96\Tstrut  \\
    \it mask-train    & 46.90                                & 43.15                                  & 0.43                                       & 4.55   & 5.85  & 11.55 & 82.61 & 0.81   & 1646    & 2.00     & 0.57      & 5.91    & 9.74         \\
    \it mask-all      & 42.53                                & 38.49                                  & 0.24                                       & 3.85   & 17.58 & 25.15 & 57.26 & 0.61   & 1677    & 1.96     & 0.61      & 5.66    & 9.16\Bstrut  \\\hdashline[0.5pt/2pt]
    \it desc-repl     & 49.35                                & 42.85                                  & 0.47                                       & 4.57   & 5.78  & 8.80  & 85.42 & 0.82   & 1693    & 1.94     & 0.59      & 5.86    & 9.18\Tstrut  \\
    \it desc-cat      & 53.07                                & 45.04                                  & 0.55                                       & 4.72   & 3.46  & 4.66  & 91.88 & 0.87   & 1668    & 1.91     & 0.59      & 5.92    & 9.11         \\
    \bottomrule
  \end{tabular}
  \caption{The summary of evaluation using automatic metrics on \textsc{Rel2text} test set. See §\ref{sec:setup} for the descriptions of the models, §\ref{sec:auto} for the descriptions of the metrics, and Table \ref{tab:autostdev} for standard deviations.}
  \label{tab:auto}
\end{table*}

\subsection{Automatic Evaluation}
\label{sec:auto}
To get a high-level overview of model behavior, we evaluate generated outputs using the GEM-metrics\footnote{\url{https://github.com/GEM-benchmark/GEM-metrics}} package \cite{gehrmann2021gem}, which provides an extensive set of automatic metrics for text generation.

\paragraph{Lexical Similarity} We first measure lexical similarity between the model outputs and human references using \textbf{BLEU} \cite{papineni2002bleu}, \mbox{\textbf{METEOR}} \cite{banerjee-lavie-2005-meteor}, and \textbf{BLEURT} \cite{sellam2020bleurt}. The first two metrics focus on n-gram overlap; the latter is a trained metric that also captures semantic similarity between the output and the reference. Although these metrics should not be used in isolation \cite{gehrmann2022repairing}, they give us a better overview of the output quality in combination with other metrics.

\paragraph{Semantic Similarity and Legibility} Lexical similarity metrics focus on the surface form, which may not be telling the whole story. For example, if the relation \mbox{\textit{parent}} denotes that $e_t$ \textit{is the parent of} $e_h$, but the entities are swapped in the generated text, the output will be incorrect, although lexical similarity metrics will be high. To get deeper insights into semantic and lexical properties of the outputs, we use NUBIA \cite{kane2020nubia}, which is a trained metric combining several features to measure ``interchangeability'' (equivalence) of two texts. The metric outputs a single score (\textbf{NB}) with a value between 0 and 1. We also report its individual underlying features:
the semantic similarity score (\textbf{SS}) on a 0-5 scale, predicted by RoBERTa \cite{liu2019roberta} finetuned on the STS-B benchmark \cite{cer-etal-2017-semeval}; the contradiction (\textbf{C}), neutral (\textbf{N}), and entailment (\textbf{E}) probabilities from RoBERTa finetuned on the MNLI challenge from the GLUE benchmark \cite{wang2018glue}; and the perplexity score (\textbf{PPL}) from vanilla GPT-2 \cite{radford2019language}, computed as a geometric mean of probabilities of the tokens in each step (this score is referenceless).

\paragraph{Lexical Diversity} To assess lexical diversity of the generated texts, we use several metrics used in previous work \cite{duvsek2020evaluating_challenge,van2018measuring}. We measure the number of unique n-grams (\textbf{U-1}), conditional entropy of bi-grams (\textbf{CE-2}), and the mean segmental type-token ratio over segment lengths of 100 (\textbf{MSTTR}; \citealp{johnson1944studies}). We also measure the average output length in tokens (\textbf{len}).

\begin{table*}[ht]
  \centering\small
  \setlength{\tabcolsep}{4pt}
  \renewcommand{\arraystretch}{1.25}
  \begin{tabular}{p{0.1cm}p{0.6cm}p{4cm}p{1.5cm}p{0cm}p{8cm}} \toprule

     & \textbf{Label} & \textbf{Example input}                                           & \textbf{Model}                            & \multicolumn{2}{p{5.75cm}}{\textbf{Example outputs (\red{\xmark} incorrect, \green{\cmark} correct)}}                                                                                                                                          \\ \midrule
    \multirow{8}{*}{\rotatebox[origin=c]{90}{\textit{model}}}
     & \textsc{Sem}   & (Yousra Matine, \emph{sport country}, Morocco)                   & \emph{mask-mask} \newline \BARTr          & \red{\xmark} \newline \green{\cmark}                                                                  & Yousra Matine was born in Morocco. \newline Yousra Matine plays for Morocco.                                                           \\
     & \textsc{Dir}   & (Kentucky Channel, \emph{former broadcast network}, KET ED)      & \mbox{\emph{fewshot-100}} \newline \BARTr & \red{\xmark} \newline \green{\cmark}                                                                  & KET ED was broadcast on Kentucky Channel ED. \newline The Kentucky Channel was broadcast on KET ED.                                    \\
     & \textsc{Lit}   & (Vietnam Television, \newline \emph{first air date}, 1970-09-07) & \BARTk \newline \BARTr                    & \red{\xmark} \newline \green{\cmark}                                                                  & The first air date of Vietnam Television was 1970-09-07. \newline Vietnam Television first aired on 1970-09-07.                        \\
     & \textsc{Lex}   & (RPG-43, \emph{used in war}, \newline The Troubles)              & \BARTr \newline \BARTk                    & \red{\xmark} \newline \green{\cmark}                                                                  & RPG-43 was used in the The Troubles. \newline The RPG-43 was used in the Troubles. \Bstrut                                             \\\hdashline[0.5pt/2pt]
    \multirow{4}{*}{\rotatebox[origin=c]{90}{\textit{data}}}
     & \textsc{Ent}   & (The Age of Entitlement, \newline \emph{by artist}, The Basics)  & \BARTk \newline \BARTr                    & \red{\xmark} \newline \green{\cmark}                                                                  & \Tstrut The Age of Entitlement was written by The Basics. \newline The Age of Entitlement was recorded by The Basics.                  \\
     & \textsc{Lbl}   & (General Motors Epsilon platform, \emph{vehicle}, Cadillac XTS)  & \BARTw \newline \emph{desc-cat}           & \red{\xmark} \newline \green{\cmark}                                                                  & General Motors Epsilon is a vehicle similar to the Cadillac XTS. \newline General Motors Epsilon platform is used in the Cadillac XTS. \\
    \bottomrule
  \end{tabular}
  \caption{Error categories used in manual analysis, with examples of errors found and corresponding correct verbalizations.
    Model error types (top):
    \textsc{Sem} -- The output is semantically incorrect,
    \textsc{Dir} -- The direction of the relation is swapped,
    \textsc{Lit} -- The verbalization is too literal,
    \textsc{Lex} -- There is a lexical error in the output.
    Input data error types (bottom):
    \textsc{Ent} -- The verbalization may depend on the entities,
    \textsc{Lbl} -- The relation label is not clear.
  }
  \label{tab:cat}
\end{table*}

\subsection{Manual Error Analysis}
\label{sec:manual}
To examine the sources of errors, we perform an in-house annotation of the model outputs.
We identify four model error types based on preliminary observations:
semantic errors (\textsc{Sem}), with a swap of the relation direction
(\textsc{Dir}) as a special case, too literal (\textsc{Lit}), i.e.\ containing awkward or misleading phrasing, and grammar/lexical errors (\textsc{Lex}). We further annotate two types of input data errors: ambiguous relations (\textsc{Ent}) and relations with unclear labels (\textsc{Lbl}).
Examples are shown in Table~\ref{tab:cat}.
We select 100 random examples together with their corresponding outputs from the \textit{\BARTr}, \textit{\BARTw}, \textit{\BARTk}, \textit{fewshot-100}, \textit{mask-all} and \textit{desc-cat} models. Without revealing the output sources, we ask three expert annotators to mark all error categories that apply.

\section{Results}
\label{sec:results}
\subsection{Automatic Evaluation Results} Table~\ref{tab:auto} shows automatic scores for all our models.
\BARTr{} is the best among the fully trained models in terms of lexical overlap metrics (which is expected, as it is trained on the most similar reference distribution), but the \BARTw{} and \BARTk{} models are almost equal in terms of semantic consistency, achieving around 90\% average entailment probability, which is on par with the copy baseline.

\begin{figure}[t]
  \centering
  \includegraphics[width=\columnwidth]{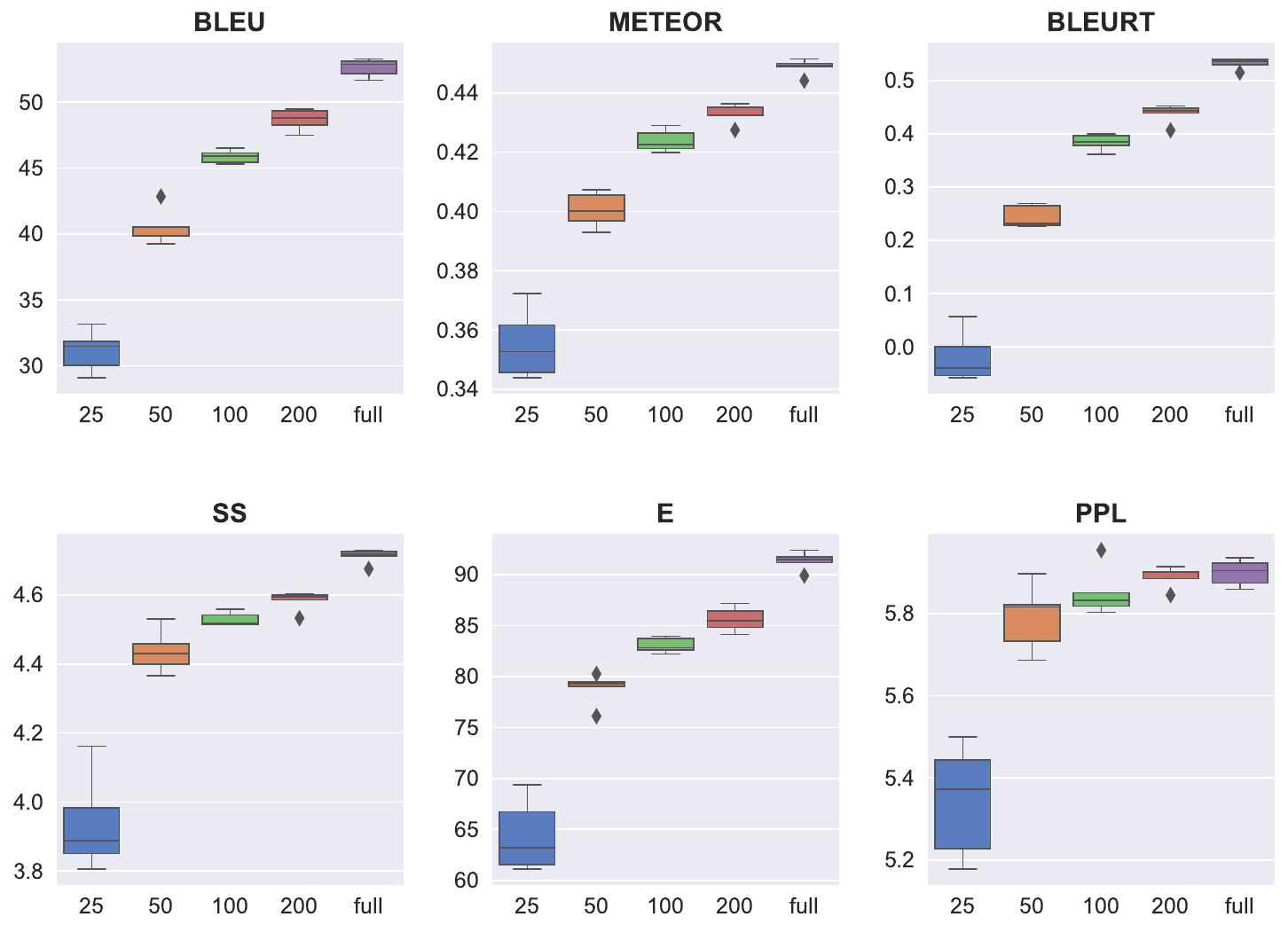}
  \caption{Boxplots for selected metrics from Table \ref{tab:auto} w.r.t. the number of examples (displayed on the \textit{x}-axis, $\textit{full} = 1522$), taking into account variance from individual random seeds (cf.\ Table~\ref{tab:autostdev}).}\label{fig:fewshot}
\end{figure}

Semantic consistency is much lower for the few-shot models (e.g.\ the average entailment probability is between 65\% and 85\%), showing that there is a certain minimum amount of data needed to achieve consistent outputs.
Using more examples for training the model generally helps to decrease variance and increase performance across various metrics  (cf.\ Figure~\ref{fig:fewshot}).

Interestingly, the models which do not see the relations during test time (\textit{mask-test} and \textit{mask-all}) still achieve around 60\% average entailment probability, similarly to the worst few-shot model. Although their rate of contradictions is higher than for other models, the results suggest that in many cases, the guessed relation is compatible with the correct relation.

Another interesting observation is that the \textit{mask-train} model (trained not to use the labels) is able to use the labels provided at test time to improve the outputs considerably (contradiction rate drops from 17\% to 5\% compared to \textit{mask-all}).
The fact that the short labels are both sufficient and necessary for the successful verbalization is emphasized by the fact that the \textit{desc-repl} model is worse than \BARTr{} (although the descriptions are longer and supposedly explain the relation semantics), and the benefits of concatenating the descriptions alongside the relation labels (\textit{desc-cat}) are negligible, only slightly improving lexical similarity metrics (0.5 BLEU point gain over \BARTr{}).

In terms of lexical diversity, human references use more unique n-grams, but the model outputs are very similar in other aspects. It remains to be seen if the model outputs can stay semantically consistent with diversity-focused decoding techniques such as nucleus sampling \cite{holtzmanBDFC20}.

\subsection{Error Analysis Results}
Results are summarized in Figure \ref{fig:manual}; complete results are presented in Appendix \ref{app:manual}.
Examples of model outputs for each error type are shown in Table~\ref{tab:cat}; more examples are given in Appendix~\ref{app:examples}.

The \BARTk{} and \BARTw{} models use expressions that are too literal (\textsc{Lit}) in 23 and 29 cases, respectively, while the \BARTr{} and \textit{desc-cat} models do the same only in 11 cases (5 out of which are marked as \textsc{Lbl}, i.e., with an unclear label). This suggests that the variability of our dataset helps models to apply more natural expressions, especially if the relation is understandable from its label.

There is a near-constant portion of examples where the models make a semantic error (\textsc{Sem}) \textit{and} the input is marked as needing an extra description (\textsc{Lbl}). The models also make relatively many semantic errors on their own, most prominently in the case of the \textit{fewshot-100} and the \textit{mask-all} models. The \textit{mask-all} model made a semantic error in 78 cases, suggesting that guessing the exact relation just from the entities is difficult (although still possible in 22 cases). Morevover, the outcomes from this model are fluent (only 4 \textsc{Lex} errors), making it hard to detect faulty cases.

The case of swapping the relation direction (\textsc{Dir}) is surprisingly not that common. This is probably down to having only a few examples in our dataset prone to this kind of error.
Notably, the results for \BARTr{} and \textit{desc-cat} are very similar, rendering the impact of extra descriptions negligible.

Finally, there were only 12 out of 100 examples annotated as \textsc{Ent}, which suggests that the verbalization of the relation can be mostly decided irrespective of the entities in the triple.

\begin{figure}[t]
  \centering
  \includegraphics[width=\columnwidth]{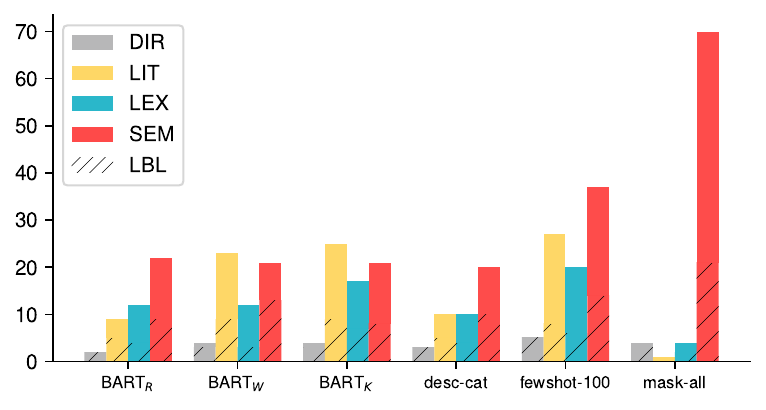}
  \caption{Number of annotated errors per model (see §\ref{sec:manual} and Table~\ref{tab:cat} for the description of error categories and §\ref{sec:experiments} for the models). The striped part signifies that the label of the input was marked as unclear. See Appendix \ref{app:manual} for details.}\label{fig:manual}
\end{figure}

\section{Downstream Tasks}
\label{sec:downstream}
Given that the \BARTr{} model can describe relations from their labels with high accuracy, we investigate if we can use the model to replace manually created templates in downstream tasks. We select two qualitatively different tasks, both using the idea of transforming individual input triples to simple sentences as a preprocessing step: tabular reasoning (§\ref{sec:tab_res}) and zero-shot data-to-text generation (§\ref{sec:zeroshot}).

\subsection{Tabular Reasoning}
\label{sec:tab_res}

\citet{gupta2020infotabs} presented the \textsc{InfoTabS} dataset as an NLI benchmark on tabular data. Each example is a structured table with a set of premises, i.e.\ natural language claims about the table; the task is to determine whether each premise is entailed by the table, contradicted by it, or neither.

They represent the table as \textit{a paragraph} where each table cell is represented as a short sentence, mostly using a simple template \text{``The \textit{key} of \textit{title} are \textit{value}.''} \citet{neeraja2021incorporating} extend \citeauthor{gupta2020infotabs}'s approach, including a \textit{better paragraph representation} for which they prepare a fine-grained set of rules for individual entity categories. The rules\footnote{Formalized using more than 250 lines of Python code: \url{https://github.com/utahnlp/knowledge\_infotabs/blob/main/scripts/preprocess/bpr.py\#L120}} aim to minimize the number of ungrammatical sentences and improve the reasoning abilities of the NLI model.

We replicate the setup of \citet{neeraja2021incorporating} for the original (OPR) and better (BPR) paragraph representation using their public codebase. We then replace their templates with our \BARTr{} model, verbalizing the triple (\textit{title}, \textit{key}, \textit{value}). The results are summarized in Table \ref{tab:nli}.

\begin{table}[t]\centering
  \small
  \setlength{\tabcolsep}{4pt}
  \begin{tabular}{lcccc}\toprule
    premise repr.                       & dev   & $\alpha_1$ & $\alpha_2$ & $\alpha_3$ \\\midrule
    OPR \cite{gupta2020infotabs}        & 76.78 & 75.30      & 68.46      & 64.63      \\
    BPR \cite{neeraja2021incorporating} & 77.04 & 74.44      & 67.46      & 63.17      \\
    \BARTr{} (ours)                     & 74.44 & 74.31      & 64.59      & 63.46      \\
    \bottomrule
  \end{tabular}
  \caption{Accuracy for the dev set and test sets  $\alpha_{1,2,3}$ from the \textsc{InfoTabS} dataset. The results are averaged over 3 random seeds.}\label{tab:nli}
\end{table}

Our preliminary manual evaluation suggests that the sentences from our model are indeed more grammatical (even compared to BPR). However, we observe that the
performance is comparable across all three test sets. In line with \citet{mccoy2019right}, we conclude that for classification tasks such as NLI, the input content appears to be more important than the input form.

\subsection{Zero-shot Data-to-Text Generation}
\label{sec:zeroshot}

\citet{kasner2022neural} proposed a setup for zero-shot D2T generation in which pretrained models are used to gradually transform text into the final description. The first step of the pipeline requires transforming individual triples into text. We focus on the WebNLG dataset, for which the authors manually created 354 templates.\footnote{Available at \url{https://github.com/kasnerz/zeroshot-d2t-pipeline/blob/main/templates/templates-webnlg.json}} We replicate the authors' setup using their public code, applying \BARTr{} instead of the templates.
The results are summarized in Table \ref{tab:zeroshot}.

\begin{table}[t]\centering
  \small
  \setlength{\tabcolsep}{4pt}
  \begin{tabular}{clcccc}\toprule
    dataset                            & model    & BLEU  & METEOR & O     & H            \\\midrule
    \multirow{2}{*}{\textit{filtered}} & orig     & 43.19 & 39.13  & 0.152 & 0.073        \\
                                       & \BARTr{} & 45.39 & 38.97  & 0.056 & 0.161\Bstrut \\\hdashline[0.5pt/2pt]
    \multirow{2}{*}{\textit{full}}     & orig     & 42.92 & 39.07  & 0.051 & 0.148\Tstrut \\
                                       & \BARTr{} & 44.63 & 38.93  & 0.058 & 0.166        \\
    \bottomrule
  \end{tabular}
  \caption{Lexical similarity metrics (BLEU, METEOR) and ommission (O) and hallucinaton (H) rate; following the setup in \citet{kasner2022neural}.}\label{tab:zeroshot}
\end{table}

We note that the pipeline using our model for preprocessing is able to achieve improvements of $\sim$2 BLEU points, at the cost of a slightly higher omission and hallucination rate, but crucially without needing the manual effort to create templates.
Cursory examination shows that sentences produced by our model are qualitatively similar to the manual templates, but more varied. Unlike the templates, our model may verbalize a relation differently depending on the context.
Overall, we argue that training a PLM on verbalizing individual relations can potentially replace the manual effort of creating simple templates, which will have a notable impact for scaling similar approaches to larger datasets.

\section{Discussion}
\label{sec:discussion}
Based on our experiments, we can conclude that PLMs are indeed able to verbalize novel relations (\textbf{RQ1}). However, there is a caveat: if the relation label is ambiguous or when the cues about the relation are limited (\textbf{RQ3}), the model will resolve to guessing and the semantic accuracy of the output descriptions may drop. A takeaway for datasets which do not follow standard naming conventions, such as the Rotowire dataset with basketball summaries \cite{wiseman2017challenges} which uses abbreviations for column headers (e.g.\ FG3A stands for \textit{``the number of shots the player attempted beyond the arc''}), is that rephrasing the labels to natural language may increase the robustness of D2T systems applied on these datasets.

We have focused on finetuned PLMs, which in our case require at least several hundreds of examples to produce satisfactory results (\textbf{RQ2}). However, recent research suggests prompting large PLMs capable of in-context learning \cite{brown2020language} may help to bring down the number of examples required close to zero \cite{li2021prefix,reynolds2021prompt,schucher2022power,chia2022relationprompt,xiang2022asdot}. In this case, the models do not have a possibility to learn the correct verbalizations from the training data, which will probably make using clear and unambiguous labels even more important: an issue to investigate in future work.

We showed that improving the outputs using longer relation descriptions is not straightforward (\textbf{RQ4}). To achieve more notable improvements, it may be necessary to combine a more detailed specification regarding the relation direction, type, acceptable values, etc., together with a model able to reason about this specification. A promising research in this direction could be using chain-of-thought reasoning, so far applied for tasks such as  open-domain question answering or solving math word problems \cite{gao2022pal,wei2022chain,yao2022react,nye2021show}.

The remaining open question is how to handle input data with noisy labels. We suggest that detecting these cases and fixing them prior to generation (for example with knowledge-augmented systems or a human-in-the-loop setup) could help to improve the robustness of D2T systems in real-world scenarios.

\section{Conclusion}
We analyzed the abilities of PLMs to verbalize unseen relations in KGs using the relation labels. Based on our findings, we believe that having expressive and unambiguous data labels is a good starting point for adapting D2T systems to new domains. For the analysis, we collected the \textsc{Rel2Text} dataset, which can help to replace the hand-crafted templates on downstream tasks. Future work may investigate how our findings generalize to prompt-based few-shot or zero-shot D2T generation with large PLMs.

\section*{Limitations}
Our analysis is limited to verbalizing single triples, which is only a stepping stone towards full-fledged G2T generation. To generate data for entire subgraphs, other issues need to be solved first, including compositional generalization and structure-aware modeling. Nevertheless, we believe that this simplified setting allows us to distill insights which are still applicable to G2T generation in general.

The factuality of the \textsc{Rel2Text} dataset is tightly related to the data in the input KGs, which may contain outdated or incorrect information, and may be influenced by our processing methods (see Appendix \ref{app:scraping} for details). Using the models trained on our dataset should be done with caution, since it can lead in producing harmful, imprecise, or factually incorrect statements.

We focus only on the English part of the KGs and English datasets. In the future, our approach could be extended to multilingual setting using multilingual PLMs and non-English parts of KGs. For more morphologically rich languages, an extra effort would have to be put into correctly inflecting the entities in the generated text.

\section*{Ethics Statement}
As we are aiming to develop D2T systems which can robustly generate text for multiple domains, we are building upon PLMs which are known to reflect or amplify biases found in their pretraining corpus \cite{bender2021dangers}. Although the purpose of our study is to minimize these biases, the outputs of our models can still contain statements which are not aligned with the input data and user needs.

We collected our training and evaluation data through the Prolific crowdsourcing platform. We ensured that all the annotators were given an average reward per hour according to the platform recommendations and we put extra attention into informing the participants about the content and purpose of our study. We also manually filtered the output to minimize the amount of noisy references in our dataset. See \ref{sec:rel2text} and Appendix \ref{app:crowdsourcing} for more details on the annotation process.

\section*{Acknowledgements}

This research was supported by Charles University projects GAUK 140320 and SVV 260575, an Apple NLU Research Grant for Heriot-Watt University and Charles University, a Carnegie Trust Research Incentive Grant (RIG009861), and by the European Research Council (Grant agreement No. 101039303 NG-NLG). It used resources provided by the LINDAT/CLARIAH-CZ Research Infrastructure (Czech Ministry of Education, Youth and Sports project No. LM2018101).

\bibliography{custom}
\bibliographystyle{acl_natbib}

\appendix

\section{Data Retrieval}
\label{app:scraping}
\paragraph*{DBPedia} We query DBPedia through its SPARQL access point: \url{http://dbpedia.org/sparql}. We retrive relations as objects of type \texttt{rdf:Property} which have a property \texttt{rdfs:comment} (i.e., the relation description) and language \texttt{'en'}.

\paragraph*{YAGO} We download the English Wikipedia subset of YAGO 4 database dump from \url{https://yago-knowledge.org/downloads/yago-4}. We retrieve all objects of type \texttt{rdf:Property} which have a property \texttt{rdfs:comment}. For the entity descriptions, we parse the entity page at YAGO website \url{http://yago-knowledge.org/resource/}.

\paragraph*{Wikidata} We first use the Wikidata SPARQL access point: \url{https://query.wikidata.org/sparql} to retrieve the list of relations as objects of type \texttt{wikibase:Property} with \texttt{wikibase:language="en"}, together with their English descriptions (\texttt{lang(?altLabel) = "en"}). Second, we query Wikidata through the LDF endpoint \url{https://query.wikidata.org/bigdata/ldf}, which is better able to handle heavy requests, to retrieve the list of triples involved in the relation. Finally, for retrieving the entity descriptions, we use the API at \url{https://www.wikidata.org/w/api.php}.

\paragraph*{Filtering} We apply a comprehensive set of filters for filtering out noisy triples, including triples with entities containing meta-information (\textit{``Category:''}, \textit{``XMLSchema\#''}), URLs, entites longer than 64 characters, relations having the string \textit{``id''}, \textit{``number''}, or \textit{``code''} in the label, or having \textit{``Reserved for DBpedia''} in the description. As a consequence, we lose some relations, most notably about 2/3 of the relations from Wikidata describing various identifiers (we opted for this step in order to maintain data diversity). If KGs contain relations with identical labels, we prefer the relations from DBPedia and YAGO (which have a substantially lower amount of relations) to Wikidata relations.

\paragraph{Missing Units} Our dataset mostly does not contain units for quantities. Although the units are usually present in the KGs, they are not part of the quantity itself -- they may be either connected to the quantity with another property, or described informally in the relation label. Since our focus was on the relation labels, we decided to not put additional effort in retrieving and processing the units. In effect, we consider verbalizations not using the units (e.g., (Bommersheim substation, \textit{voltage}, 20000) $\rightarrow$ ``Bommersheim substation has a voltage of 20000.'') as correct.

\paragraph*{Factual Correctness} A certain part of the data is factually incorrect, either because there was an error in the knowledge graph (e.g., (Catalans, \textit{population place}, \textbf{Italy})) or because there was a processing error (e.g., (Child Language Teaching and Therapy, \textit{final publication year}, \textbf{-1985)}. Since our focus was not on judging the factuality of the inputs (which is a difficult problem on its own right), we decided to keep the examples in the dataset and consider the examples semantically consistent with the input triple as correct.

\paragraph*{Other Notes} \begin{itemize}
  \item All the data was retrieved in February 2022, except for YAGO where we used the newest available dump \textit{2020-02-24}.
  \item Although we retrieved the entity descriptions wherever possible and we include them in our dataset, we decided not to use them in our experiments.
  \item The Python code for retriving the data is available in the paper repository.
\end{itemize}

\section{Crowdsourcing Details}
\label{app:crowdsourcing}

\begin{figure}[ht!]
  \centering
  \includegraphics[width=\columnwidth]{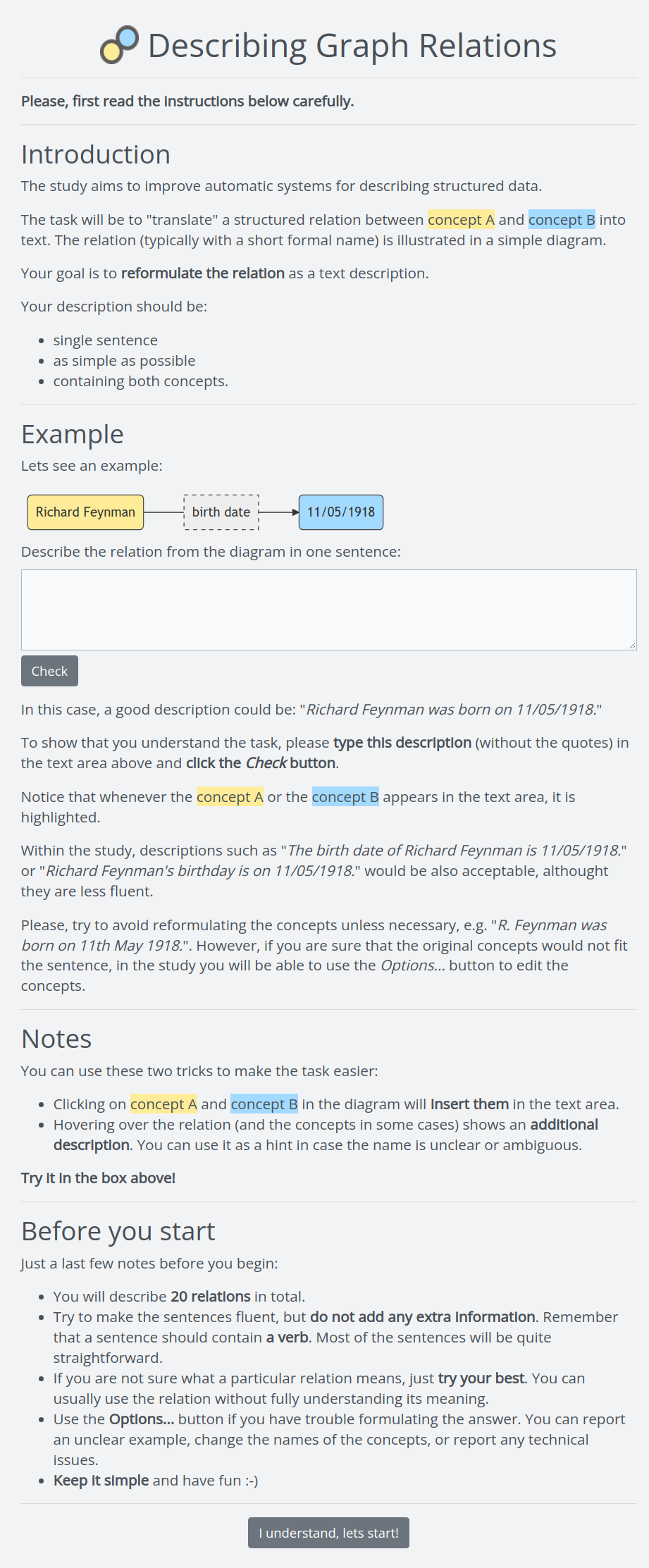}

  \caption{The introduction screen shown to the participants.}\label{fig:intro}
\end{figure}

\begin{figure}[ht!]
  \centering
  \includegraphics[width=\columnwidth]{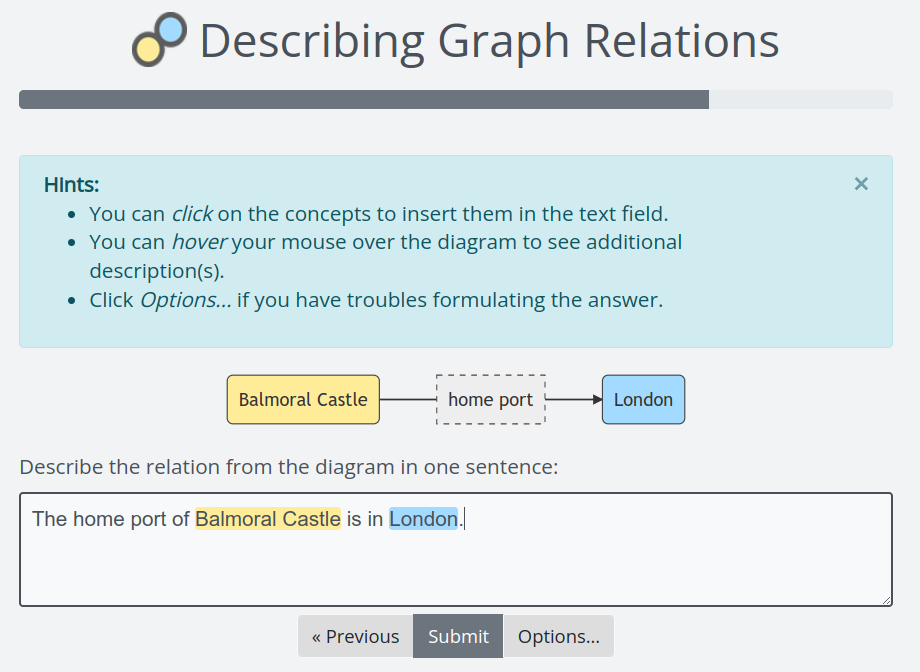}

  \caption{The annotation inteface.}\label{fig:ann}
\end{figure}

We built a web interface for collecting verbalizations for the triples. Figure \ref{fig:intro} shows the introductory instructions displayed for the participants and Figure \ref{fig:ann} shows the annotation interface.

We hired annotators on the Prolific crowdsourcing platform \url{https://app.prolific.co/}. We required that the annotators are native speakers of English. After completing an introductory example, the annotators were given 20 randomly selected triples presented in a sequential order. The annotators were asked to write a short, single-sentence description of the triple. For making the annotation easier, hovering the mouse over the relation revealed its description (this applied also for the entities, if the description was present).

The annotators could also click on the entity to insert it in the text. This motivated the users to insert the entities in the original form. Once the entity appeared in the text (either typed or inserted), it was highlighted. We required that both entities (and at least two extra characters) are present in the text before proceeding to the next step. Because of this requirement, approximately 98.6\% sentences in our dataset can be delexicalized using exact string matching. The users also had an option to modify the entity name, which would be recorded as a new ground-truth input (e.g., to make its form more natural). However, this option was used only sparingly.

In total, we collected 8,265 responses for 7,334 examples. Multiple responses for some examples are a consequence of random selection combined with sessions running in parallel. In the final dataset used in our experiments, we selected at most one correct answer for each example (see Appendix \ref{app:postprocessing}).

\section{Postprocessing the Dataset}
\label{app:postprocessing}
Two of the paper authors manually postprocessed the dataset. We used the following criteria for marking the responses:

\begin{itemize}
  \item \textbf{OK} -- The sentence is fluent and semantically consistent with the input.
  \item \textbf{Noisy} -- The sentence contains a minor typographical or grammatical error, or the sentence sounds ``awkward'' (e.g., the relation label is used too literally).
  \item \textbf{Corrupted} -- The sentence is semantically incorrect, contains a major typographical or grammatical error, or generally does not make sense.
  \item \textbf{Extra information} -- The sentence is correct, but contains extra information about the entities which cannot be derived from the triple itself (e.g., the country of origin of the person found in the entity description).
\end{itemize}

\begin{figure}[ht!]
  \centering
  \includegraphics[width=\columnwidth]{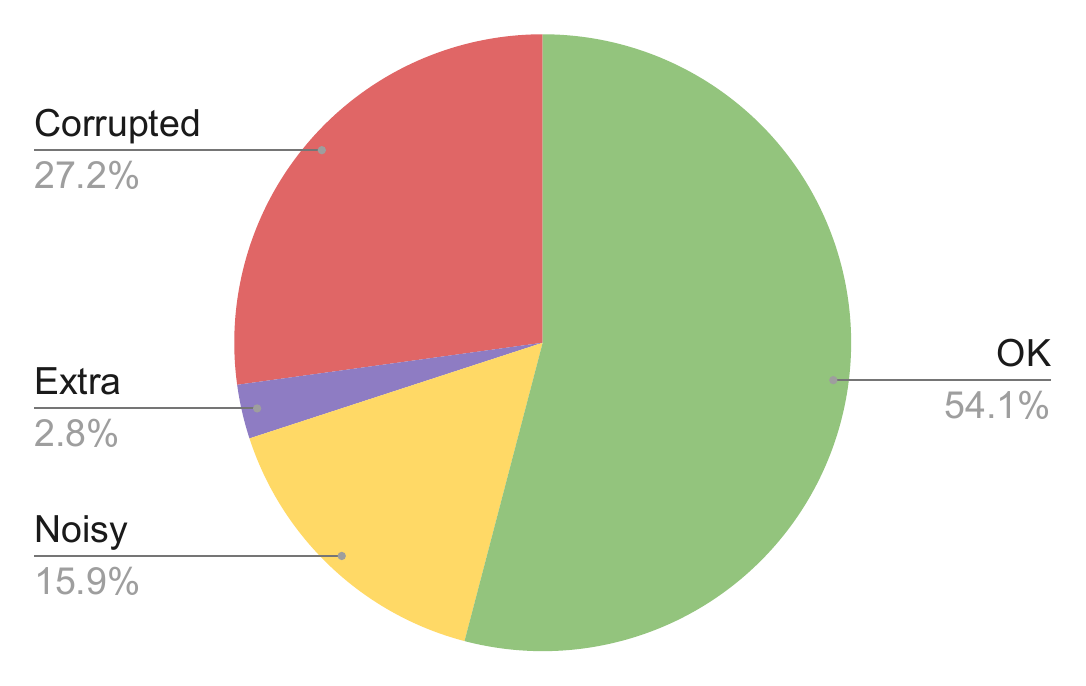}
  \caption{The distribution of crowdsourced responses in our dataset.}\label{fig:chart}
\end{figure}

Figure \ref{fig:chart} shows the distribution of responses in our dataset. We marked 4,469 (54.1\%) responses as \textit{OK}, 1,314 (15.9\%) responses as \textit{Noisy}, 2,246 (27.2\%) responses as \textit{Corrupted} and 235 (2.8\%) responses as \textit{Extra information}.

Because our priority was to have clean data for evaluation, we decided to use only the \textit{OK} part of our dataset in our experiments. We only use one example for each input triple in our experiments, which gives 4,097 instances. However, since we believe that the human outputs can also be an interesting research target, e.g. for investigating the feasibility of verbalizing the input data, we release all the annotations for future investigations.

\begin{table*}[htb!]\centering
  \footnotesize
  \begin{tabular}{lc>{\hspace{-2mm}}c>{\hspace{-2mm}}c>{\hspace{-2mm}}ccccccc>{\hspace{-2mm}}c>{\hspace{-2mm}}c>{\hspace{-1mm}}cc}\toprule
    \multirow{2}{*}{ \textbf{experiments}} & \multicolumn{2}{c}{ \textbf{Lexical}} & \multicolumn{6}{c}{ \textbf{Semantics}} & \multicolumn{5}{c}{ \textbf{Referenceless}}                                                                                                      \\\cmidrule{2-14}
                                           & \bf BLEU                              & \bf METEOR                              & \bf BLEURT                                  & \bf SS & \bf C & \bf N & \bf E & \bf NB & \bf U-1 & \bf CE-2 & \bf MSTTR & \bf PPL & \bf len       \\\midrule
    \it \BARTr{}                           & 0.60                                  & 0.30                                    & 0.01                                        & 0.02   & 0.41  & 0.38  & 0.65  & 0.01   & 7       & 0.07     & 0.01      & 0.03    & 0.10 \Tstrut  \\
    \it \BARTw{}                           & 0.69                                  & 0.09                                    & 0.00                                        & 0.02   & 0.23  & 0.94  & 1.07  & 0.00   & 7       & 0.02     & 0.00      & 0.02    & 0.10          \\
    \it \BARTk{}                           & 0.78                                  & 0.22                                    & 0.01                                        & 0.02   & 0.49  & 0.15  & 0.42  & 0.01   & 11      & 0.03     & 0.00      & 0.03    & 0.06 \Bstrut  \\\hdashline[0.5pt/2pt]
    \it fewshot-25                         & 1.60                                  & 1.18                                    & 0.05                                        & 0.14   & 1.19  & 2.67  & 3.58  & 0.03   & 68      & 0.05     & 0.02      & 0.14    & 0.61\Tstrut   \\
    \it fewshot-50                         & 1.36                                  & 0.59                                    & 0.02                                        & 0.06   & 0.99  & 0.77  & 1.59  & 0.02   & 19      & 0.13     & 0.01      & 0.08    & 0.19          \\
    \it fewshot-100                        & 0.51                                  & 0.38                                    & 0.02                                        & 0.02   & 0.63  & 0.53  & 0.75  & 0.01   & 14      & 0.11     & 0.01      & 0.06    & 0.25          \\
    \it fewshot-200                        & 0.80                                  & 0.35                                    & 0.02                                        & 0.02   & 0.60  & 1.37  & 1.25  & 0.01   & 14      & 0.06     & 0.00      & 0.02    & 0.08  \Bstrut \\\hdashline[0.5pt/2pt]
    \it mask-test                          & 0.25                                  & 0.11                                    & 0.01                                        & 0.01   & 0.62  & 0.48  & 1.00  & 0.01   & 10      & 0.03     & 0.00      & 0.03    & 0.06\Tstrut   \\
    \it mask-train                         & 0.19                                  & 0.09                                    & 0.01                                        & 0.03   & 0.64  & 1.73  & 1.95  & 0.01   & 10      & 0.02     & 0.01      & 0.03    & 0.09          \\
    \it mask-all                           & 1.19                                  & 0.22                                    & 0.00                                        & 0.04   & 1.29  & 0.97  & 1.62  & 0.01   & 8       & 0.03     & 0.01      & 0.05    & 0.19 \Bstrut  \\\hdashline[0.5pt/2pt]
    \it desc-repl                          & 0.29                                  & 0.13                                    & 0.00                                        & 0.01   & 0.71  & 0.51  & 0.40  & 0.01   & 10      & 0.05     & 0.00      & 0.03    & 0.14\Tstrut   \\
    \it desc-cat                           & 0.57                                  & 0.21                                    & 0.00                                        & 0.01   & 0.24  & 0.28  & 0.42  & 0.00   & 10      & 0.04     & 0.01      & 0.03    & 0.09          \\
    \bottomrule
  \end{tabular}
  \caption{Standard deviations for the model experiments in Table \ref{tab:auto}. Results were computed for 5 random seeds.}
  \label{tab:autostdev}
\end{table*}

\paragraph{Sidenote: Entity Overlap} The overlap between entities in test set of Rel2Text and train sets of Rel2Text and WebNLG is around 1\%, only 0.5\% being named entities (the rest are numerical values). On the contrary, around half of the entities in Rel2Text test set are included in the KeLM train set since KeLM covers a large portion of Wikidata with ca. 6M unique entities. In general, we believe that entity overlap does \textit{not} have a notable influence on the results since all the pretrained models already have a representation of the entities from the pretraining stage.

\section{Experimental Setup}
\label{app:hyperparams}
\paragraph*{Framework} We implemented the models in PyTorch Lightning \cite{paszke2019pytorch}. We used the PyTorch \cite{falcon2019pytorch} version of \textsc{BART-base} from the Huggingface library \cite{wolf2019huggingface}, with 140M parameters as a basis for all our models.

\paragraph*{Hyperparameters} We use the Adam \cite{kingmaB14} optimizer ($\beta_1=0.9, \beta_2=0.98, \varepsilon=1^{-6} $) with learning rate $2^{-5}$ and polynomial scheduling with 10\% warmup steps. We train the models with batches of size 8 and accumulating gradients with factor 4 (an effective batch size of 32).

\paragraph{Training} We train the models for 10 epochs on a single GeForce RTX 3090 GPU with 24 GB RAM, except for \BARTk{} model which we train for 1 epoch. Training times were around 15 minutes for the datasets based on \textsc{Rel2Text}, 2 hours for \BARTw{} and 3 days for \BARTk{}.  We use greedy decoding in all our experiments.

\section{Automatic Evaluation}

The standard deviations for each experiment from Table \ref{tab:auto} are listed in Table \ref{tab:autostdev}.
\section{Manual Evaluation}
\label{app:manual}

Table \ref{tab:manual} shows full results of our manual evaluation.

\begin{table}[ht!]
  \centering\footnotesize
  \begin{tabular}{l ccccc } \toprule
                         & \textsc{dir}           & \textsc{lit}            & \textsc{lex}            & \textsc{sem}             \\\midrule
    \textit{\BARTr{}}    & 2 \scriptsize{(0,2,0)} & 11 \scriptsize{(1,5,1)} & 12 \scriptsize{(0,4,0)} & 24 \scriptsize{(2,9,1)}  \\
    \textit{\BARTw{}}    & 8 \scriptsize{(2,3,2)} & 23 \scriptsize{(2,9,0)} & 12 \scriptsize{(1,3,0)} & 25 \scriptsize{(4,13,2)} \\
    \textit{\BARTk{}}    & 6 \scriptsize{(1,3,1)} & 29 \scriptsize{(3,9,2)} & 19 \scriptsize{(1,7,1)} & 25 \scriptsize{(3,8,2)}  \\
    \textit{fewshot-100} & 3 \scriptsize{(0,3,0)} & 12 \scriptsize{(2,5,1)} & 10 \scriptsize{(0,4,0)} & 24 \scriptsize{(2,10,2)} \\
    \textit{desc-cat}    & 5 \scriptsize{(0,5,0)} & 31 \scriptsize{(3,8,2)} & 22 \scriptsize{(2,6,1)} & 41 \scriptsize{(6,14,2)} \\
    \textit{mask-all}    & 4 \scriptsize{(1,3,0)} & 1 \scriptsize{(1,0,0)}  & 4 \scriptsize{(1,2,0)}  & 78 \scriptsize{(7,21,4)} \\\bottomrule
  \end{tabular}\caption{Number of errors manually annotated in 100 outputs from the models. The parentheses contain the portion of examples where the input was marked as \textsc{Ent} only, \textsc{Lbl} only, and \textsc{ENT+DESC}, respectively.}
  \label{tab:manual}
\end{table}

In total, there were 7 examples annotated as \textsc{Ent} only, 24 examples annotated as \textsc{Lbl} only, and 5 examples annotated as \textsc{ENT+DESC}.

\section{Examples}
\label{app:examples}
Examples of outputs of the models selected for manual evaluation are shown in  Table \ref{tab:examples}.

\begin{table*}[htbp]\centering
  \begin{tabular}{l p{13.5cm}} \toprule
    \textbf{input}          & (Kalatungan Mountain Range, \textit{highest}, Mount Kalatungan)                    \\
    \textbf{desc.}          & \textit{The highest mountain of a mountain range.}
    \Bstrut                                                                                                      \\\hdashline[0.5pt/2pt]
    \textit{ full-rel2text} & Kalatungan Mountain Range's highest mountain is Mount Kalatunan.\Tstrut            \\
    \textit{full-webnlg}    & The Kalatungan Mountain Range is the highest of Mount Kalatuna.                    \\
    \textit{full-kelm}      & The highest mountain in the Kalatungan Mountain Range is Mount Kalatunagan.        \\
    \textit{desc-cat}       & Kalatungan Mountain Range is the highest mountain in the Kalatunan mountain range. \\
    \textit{fewshot-100}    & Kalatungan Mountain Range is the highest mountain in the Kalatunan Mountain Range. \\
    \textit{mask-all}       & Kalatungan Mountain Range is located on Mount Kalatunan.                           \\
  \end{tabular}
  \begin{tabular}{l p{13.5cm}} \toprule
    \textbf{input}          & (The Fight Before Christmas, \textit{part of series}, The Simpsons)                    \\
    \textbf{desc.}          & \textit{The series to which this episode or season belongs.} \Bstrut                   \\\hdashline[0.5pt/2pt]
    \textit{ full-rel2text} & The Fight Before Christmas is part of The Simpsons.\Tstrut                             \\
    \textit{full-webnlg}    & The Simpsons is the series that has The Fight Before Christmas as one of its episodes. \\
    \textit{full-kelm}      & The Fight Before Christmas is part of the Simpsons series.                             \\
    \textit{desc-cat}       & The Fight Before Christmas is a part of The Simpsons.                                  \\
    \textit{fewshot-100}    & The Fight Before Christmas was part of series of The Simpsons.                         \\
    \textit{mask-all}       & The Fight Before Christmas was a parody of The Simpsons.                               \\
  \end{tabular}
  \begin{tabular}{l p{13.5cm}} \toprule
    \textbf{input}          & (Landwasser Viaduct, \textit{construction material}, Limestone)                         \\
    \textbf{desc.}          & \textit{Construction material (eg. concrete, steel, iron, stone, brick, wood).} \Bstrut \\\hdashline[0.5pt/2pt]
    \textit{ full-rel2text} & Landwasser Viaduct was built using Limestone.\Tstrut                                    \\
    \textit{full-webnlg}    & The construction material of the Landwasser Viaduct is Limestone.                       \\
    \textit{full-kelm}      & The Landwasser Viaduct is made of limestone.                                            \\
    \textit{desc-cat}       & Landwasser Viaduct was built using Limestone.                                           \\
    \textit{fewshot-100}    & Landwasser Viaduct was built using Limestone.                                           \\
    \textit{mask-all}       & Landwasser Viaduct crosses Limestone.                                                   \\
  \end{tabular}
  \begin{tabular}{l p{13.5cm}} \toprule
    \textbf{input}          & (Sharnbrook Academy, \textit{county}, Bedfordshire)     \\
    \textbf{desc.}          & \textit{The county where the thing is located.} \Bstrut \\\hdashline[0.5pt/2pt]
    \textit{ full-rel2text} & Sharnbrook Academy is in Bedfordshire.\Tstrut           \\
    \textit{full-webnlg}    & Sharnbrook Academy is in the county of Bedfordshire.    \\
    \textit{full-kelm}      & Sharnbrook Academy is located in Bedfordshire.          \\
    \textit{desc-cat}       & Sharnbrook Academy is in Bedfordshire.                  \\
    \textit{fewshot-100}    & Sharnbrook Academy is in Bedfordshire.                  \\
    \textit{mask-all}       & Sharnbrook Academy is in Bedfordshire.                  \\
  \end{tabular}
  \begin{tabular}{l p{13.5cm}} \toprule
    \textbf{input}          & (Loco-Motion, \textit{computing platform}, Tomy Tutor)                                                      \\
    \textbf{desc.}          & \textit{some sort of hardware architecture or software framework, that allows this software to run} \Bstrut \\\hdashline[0.5pt/2pt]
    \textit{ full-rel2text} & Loco-Motion is run on Tomy Tutor.\Tstrut                                                                    \\
    \textit{full-webnlg}    & Tomy Tutor is the computing platform for Loco-Motion.                                                       \\
    \textit{full-kelm}      & Loco-Motion is available for Tomy Tutor.                                                                    \\
    \textit{desc-cat}       & Loco-Motion runs on Tomy Tutor.                                                                             \\
    \textit{fewshot-100}    & Loco-Motion is a computing platform for Tomy Tutor.                                                         \\
    \textit{mask-all}       & Loco-Motion was inspired by Tomy Tutor.                                                                     \\  \bottomrule
  \end{tabular}

  \caption{Examples of outputs of the models selected for manual evaluation.}
  \label{tab:examples}
\end{table*}

\end{document}